\documentclass[11pt,a4paper]{article}
\usepackage[hyperref]{naaclhlt2018}
\usepackage{times}
\usepackage{latexsym}

\usepackage{times}
\usepackage{latexsym}
\usepackage{tabularx}
\usepackage[justification=centering]{caption}
\usepackage{booktabs}
\usepackage{graphicx}
\usepackage{url}
\usepackage{enumitem}
\usepackage[symbol]{footmisc}
\usepackage{footnote}

\usepackage{url}
\aclfinalcopy

\title{The Limitations of Cross-language Word Embeddings Evaluation}

\author{Amir Bakarov\textsuperscript{\textdagger*} \quad Roman Suvorov\textsuperscript{*} \quad Ilya Sochenkov\textsuperscript{$\ddagger$*} \\
  \textsuperscript{\textdagger}National Research University Higher School of Economics, \\
  \textsuperscript{*}Federal Research Center `Computer Science and Control' of the Russian Academy of Sciences, \\
  \textsuperscript{$\ddagger$}Skolkovo Institute of Science and Technology (Skoltech), \\
  Moscow, Russia \\
  {\tt amirbakarov@gmail.com, rsuvorov@isa.ru, ivsochenkov@gmail.com} }

\date{}

\begin{document}
\maketitle
\begin{abstract}
The aim of this work is to explore the possible limitations of existing methods of cross-language word embeddings evaluation, addressing the lack of correlation between intrinsic and extrinsic cross-language evaluation methods. To prove this hypothesis, we construct English-Russian datasets for extrinsic and intrinsic evaluation tasks and compare performances of 5 different cross-language models on them. The results say that the scores even on different intrinsic benchmarks do not correlate to each other. We can conclude that the use of human references as ground truth for cross-language word embeddings is not proper unless one does not understand how do native speakers process semantics in their cognition.
\end{abstract}

\section{Introduction}
Real-valued word representations called \textit{word embeddings} are an ubiquitous and effective technique of semantic modeling. So it is not surprising that cross-language extensions of such models (\textit{cross-language word embeddings}) rapidly gained popularity in the NLP community \cite{vulic2013cross}, proving their effectiveness in certain cross-language NLP tasks \cite{upadhyay2016cross}. However, the problem of proper evaluation of any type of word embeddings still remains open.

In recent years there was a critique to mainstream methods of intrinsic evaluation: some researchers addressed subjectivity of human assessments, obscurity of instructions for certain tasks and terminology confusions \cite{faruqui2016problems,batchkarov2016critique}. Despite all these limitations, some of the criticized methods (like the word similarity task) has been started to be actively applied yet for cross-language word embeddings evaluation \cite{camacho2017semeval,camacho2015framework}. 

We argue that if certain tasks are considered as not proper enough for mono-lingual evaluation, then it should be even more inappropriate to use them for cross-language evaluation since new problems would appear due to the new features of cross-linguality wherein the old limitations still remain. Moreover, it is still unknown for the field of cross-language word embeddings, are we able to make relevant predictions on performance of the model on one method, using another. We do not know whether can we use the relative ordering of different embeddings obtained by evaluation on an intrinsic task to decide which model will be better on a certain extrinsic task. So, the aim of this work is to highlight the limitations of cross-language intrinsic benchmarks, studying the connection of outcomes from different cross-language word embeddings evaluation schemes (intrinsic evaluation and extrinsic evaluation), and explain this connection by addressing certain issues of intrinsic benchmarks that hamper us to have a correlation between two evaluation schemes. In this study as an extrinsic task we consider the cross-language paraphrase detection task. This is because we think that the model's features that word similarity and paraphrase detection evaluate are very close: both of them test the quality of semantic modeling (i.e. not the ability of the model to identify POS tags, or the ability to cluster words in groups, or something else) in terms of properness of distances in words pairs with certain types of semantic relations (particularly, semantic similarity). Therefore, we could not say that a strong difference in performances of word embeddings on these two tasks could be highly expected.

In this paper we propose a comparison of 5 cross-language models on extrinsic and intrinsic datasets for English-Russian language pair constructed specially for this study. We consider Russian because we are native speakers of this language (hence, we are able to adequately construct novel datasets according the limitations that we address). 

Our work is a step towards exploration of the limitations of cross-language evaluation of word embeddings, and it has three primary contributions:
\begin{enumerate}
\item We propose an overview of limitations of current intrinsic cross-language word embeddings evaluation techniques;
\item We construct 12 cross-language datasets for evaluation on the word similarity task;
\item We propose a novel task for cross-language extrinsic evaluation that was never addressed before from the benchmarking perspective, and we create a human-assessed dataset for this task.
\end{enumerate}

This paper is organized as follows. Section 2 puts our work in the context of previous studies. Section 3 describes the problems of intrinsic cross-language evaluation. Section 4 is about the experimental setup. The results of the comparison are reported in Section 5, while Section 6 concludes the paper.

\section{Related Work}

First investigation of tasks for cross-language word embeddings evaluation was proposed in 2015 \cite{camacho2015framework}. This work was the first towards mentioning the problem of lack of lexical one-to-one correspondence across different languages from the evaluation perspective. However, no detailed insights on limitations of evaluation (e.g. effect of this lack on evaluation scores) was reported. 2015 also saw an exploration of the effect of assessments' language and the difference in word similarity scores for different languages \cite{leviant2015separated}.

In 2016 the first survey of cross-language intrinsic and extrinsic evaluation techniques was proposed \cite{upadhyay2016cross}. The results of this study did not address the correlation of intrinsic evaluation scores with extrinsic ones (despite that the lack of correlation of intrinsic and extrinsic tasks for mono-language evaluation was proved \cite{schnabel2015evaluation}, it is not obvious if this would also extend to cross-language evaluation). In 2017 a more extensive overview of cross-language word embeddings evaluation methods was proposed \cite{ruder2017survey}, but this study did not considered any empirical analysis. 

After all, we are aware of certain works on a topic of cross-language evaluation from the cross-language information retrieval community \cite{braschler2000evaluation}, but there are no works that highlight non-trivial issues of cross-language systems evaluation from the position of word embeddings.  

\section{Problems of Cross-language Evaluation}

We address the following problems that could appear on any kind of evaluation of cross-language word embeddings against human references on any intrinsic task:
\begin{enumerate}
\item \textbf{Translation Disagreement}. Some researchers have already faced the limitations of machine word translation for constructing cross-language evaluation datasets from mono-language ones by translating them word-by-word. The obtained problems were in two different words with the same translation or with different parts of speech \cite{camacho2015framework}. We also argue that some words could have no translations while some words could have multiple translations. Of course, these issues could be partially avoided if the datasets would be translated manually and the problematic words would be dropped from the cross-language dataset, but it is not clear how the agreement for word dropping of human assessors could be concluded.  
\item \textbf{Scores Re-assessment}. Some researchers obtain new scores reporting human references by automatically averaging the scores from the mono-language datasets of which the new dataset is constructed. Another option of scores re-assessment proposes manual scoring of a new dataset by bilingual assessors. We consider that both variants are not proper since it is unclear how the scores in the cross-language dataset should be assessed: humans usually do not try to identify a similarity score between word $a$ in language $A$ and word $b$ in language $B$ since of difference in perception of these words in cognition of speakers of different languages.
\item \textbf{Semantic Fields}. According to the theories of \textit{lexical typology}, the meaning of a properly translated word could denote a bit different things in a new languages. Such effect is called \textit{semantic shift}, and there is a possibility that the actual meanings of two corresponding words could be different even if they are correctly translated and re-assessed \cite{ryzhova2016typology}. One of the ways of avoiding this problem is to exclude \textit{relational nouns} which are words with non-zero valency \cite{koptjevskaja2015semantics} from the dataset, so it should consist only of zero valency nouns that are more properly linked with real world objects. However, the distinction of words on relational and non-relational ones is fuzzy, and such assessments could be very subjective (also, since verbs are usually highly relational, they should not be used in cross-language evaluation).
\item \textbf{New Factors for Bias}. It is already known that existence of \textit{connotative associations} for certain words in mono-language datasets could introduce additional subjectivity in the human assessments \cite{liza2016improved}. We argue that yet more factors could be the cause of assessors' bias in the cross-language datasets. For example, words \textit{five} and \textit{clock} could be closely connected in minds of English speakers (since of the common \textit{five o'clock tea} collocation), but not in minds of speakers of other languages, and we think that a native English speaker could assess biased word similarity scores for this word pair.    
\end{enumerate}

\begin{table*}[t]
\begin{center}
\begin{tabular}{l|ccccc}
\toprule 
    & MSE & MUE & VM & BCCA & MFT \\
\midrule
 S.RareWord-958 (56.3\%) \cite{luong2013better} & \textbf{0.44} & 0.42 & 0.43 & 0.43 & 0.43 \\
 S.SimLex-739 (95.9\%) \cite{hill2016simlex} & 0.34 & 0.32 & \textbf{0.35} & 0.34 & 0.34 \\
 S.SemEval-243 (88.0\%) \cite{camacho2017semeval} & \textbf{0.6} & 0.56 & 0.35 & 0.34 & 0.34 \\
 S.WordSim-193 (96.4\%) \cite{agirre2009study} & 0.69 & 0.67 & \textbf{0.72} & 0.67 & 0.71 \\
 S.RG-54 (83.1\%) \cite{rubenstein1965contextual} & \textbf{0.68} & 0.67 & 0.63 & 0.61 & 0.61 \\
  S.MC-28 (93.3\%) \cite{miller1991contextual} & 0.66 & 0.7 & 0.71 & \textbf{0.72} & 0.7 \\
\midrule
  V. SimVerb-3074 (87.8\%) \cite{gerz2016simverb} & 0.2 & 0.2 & \textbf{0.23} & 0.22 & 0.21 \\
   V.Verb-115 (85.4\%) \cite{baker2014unsupervised} & 0.24 & \textbf{0.39} & 0.27 & 0.27 & 0.27 \\
 V.YP-111 (88.5\%) \cite{yang2006verb} & 0.22 & \textbf{0.37} & 0.25 & 0.25 & 0.25 \\
\midrule
 R.MEN-1146 (94.7\%) \cite{bruni2014multimodal} & 0.68 & 0.66 & \textbf{0.69} & 0.66 & 0.68 \\
 R.MTurk-551 (91.7\%) \cite{halawi2012large} & 0.56 & 0.51 & \textbf{0.57} & 0.54 & \textbf{0.57} \\
 R.WordSim-193 (96.4\%) \cite{agirre2009study} & 0.55 & 0.53 & \textbf{0.57} & 0.53 & 0.55 \\
 \midrule
 P@1, dictionary induction & 0.31 & 0.16 & \textbf{0.32} & 0.29 & 0.21 \\
 P@5, dictionary induction & \textbf{0.53} & 0.34 & 0.52 & 0.49 & 0.38 \\
 P@10, dictionary induction & \textbf{0.61} & 0.42 & 0.5 & 0.55 & 0.45 \\
\midrule
 F1, paraphrase detection, \textit{our dataset} & 0.82 & 0.77 & 0.84 & 0.83 & \textbf{0.86} \\
 F1, paraphrase detection, \textit{parallel sentences} & 0.55 & 0.45 & 0.57 & \textbf{0.6} & 0.59 \\
    \bottomrule
\end{tabular}
\end{center}
\caption{Performance of the compared models across different tasks. Evaluation on first 11 datasets indicate Spearman's rank correlation. For word similarity task: words before the hyphen in datasets name report the name of the original English dataset, the number after the hyphen report the amount of word pairs, the numbers in brackets report ratio to its English original and the prefix before the dot in the name report type of assessments.}
\end{table*}

\section{Experimental Setup}

\subsection{Distributional Models}

To propose a comparison, we used 5 cross-language embedding models.
\begin{enumerate}
\item \textbf{MSE} (\textit{Multilingual Supervised Embeddings}). Trains using a bilingual dictionary and learns a mapping from the source to the target space using Procrustes alignment \cite{conneau2017word}. 
\item \textbf{MUE} (\textit{Multilingual Unsupervised Embeddings}). Trains learning a mapping from the source to the target space using adversarial training and Procrustes refinement \cite{conneau2017word}.
\item \textbf{VecMap}. Maps the source into the target space using a bilingual dictionary or shared numerals minimizing the squared Euclidean distance between embedding matrices \cite{artetxe2018generalizing}. 
\item \textbf{BiCCA} (\textit{Bilingual Canonical Correlation Analysis}). Projects vectors of two different languages in the same space using CCA \cite{faruqui-dyer:2014:EACL}.
\item \textbf{MFT} (\textit{Multilingual FastText}). Uses SVD to learn a linear transformation, which aligns monolingual vectors from two languages in a single vector space \cite{smith2017offline}.
\end{enumerate}

We mapped vector spaces of Russian and English \textit{FastText} models trained on a dump of Wikipedia \cite{bojanowski2016enriching} with an English-Russian bilingual dictionary \cite{conneau2017word} (only one translation for a single word). 

\subsection{Intrinsic Tasks}

\textbf{Word Semantic Similarity.} The task is to predict the similarity score for a word $a$ in language $A$ and a word $b$ in language $B$. All three publicly available datasets for cross-language word similarity \cite{camacho2015framework, camacho2017semeval} are not available for Russian, so we created the cross-language datasets ourselves. We used 5 English datasets assessed by \textit{semantic similarity} of nouns and adjectives (\textit{S}), 3 datasets assessed by \textit{semantic similarity} of verbs (\textit{V}), and 3 datasets assessed by \textit{semantic relatedness} of nouns and adjectives (\textit{R}); we labeled each with a letter reporting the type of relations. We translated these datasets, merged into cross-language sets (the first word of each word pair was English, and the second was Russian), dropped certain words pairs according to limitations addressed by us (in the Section 2), and re-assessed the obtained cross-languages datasets with the help of 3 English-Russian volunteers, having Krippendorff's alpha 0.5 (final amount of word pairs and ratio to original datasets is reported at Table 1). Then we compared human references of these datasets with cosine distances of cross-language word vectors, and computed Spearman's rank correlation coefficient ($p-value$ in all cases was lower than 0.05).

\textbf{Dictionary Induction} (also called \textit{word translation}). The second task is to translate a word in language $A$ into language $B$, so for the seed word the model generates a list of the closest word in other language, and we need to find the correct translation in it. As a source of correct translations we used English-Russian dictionary of 53 186 translation pairs \cite{conneau2017word}. The evaluation on this measure was proposed as a precision on $k$ nearest vectors of a word embedding model for $k = {1, 5, 10}$. 

\subsection{Extrinsic Task and Our Dataset}

\textbf{Cross-language Paraphrase Detection}. In an analogy with a monolingual paraphrase detection task (also called \textit{sentence similarity identification}) \cite{androutsopoulos2010survey}, the task is to identify whether sentence $a$ in language $A$ and sentence $b$ in language $B$ are paraphrases or not. This task is highly scalable, and usually figures as a sub-task of bigger tasks like cross-language plagiarism detection.

We are not aware of any dataset for this task, so we designed a  benchmark ourselves for English-Russian language pair. The dataset was constructed on the base of Wikipedia articles covering wide range of topics from technology to sports. It contains 8 334 sentences with a balanced class distribution. The assessments and translations were done by 3 bilingual assessors.  The negative results were obtained by automatically randomly sampling another sentence in the same domain from the datasets.

Translations were produced manually by a pool of human translators. Translators could paraphrase the translations using different techniques (according to our guidelines), and the assessors had to verify paraphrase technique labels and annotate similarity of English-Russian sentences in binary labels. We invited 3 assessors to estimate inter-annotator agreement. To obtain the evaluation scores, we conducted 3-fold cross validation and trained Logistic Regression with only one feature: cosine similarity of two sentence vectors. Sentence representations were built by averaging their word vectors. 

In order to validate the correctness of results on our dataset, we automatically constructed a paraphrase set from a corpus of 1 million English-Russian parallel sentences from WMT'16\footnote{\path{https://translate.yandex.ru/corpus}}, generating for each sentence pair a semantic negative sample, searching for nearest sentence with a monolingual FastText model.  

\section{Results and Discussion}

\begin{figure}
	\centering
    \includegraphics[width=0.35\textwidth]{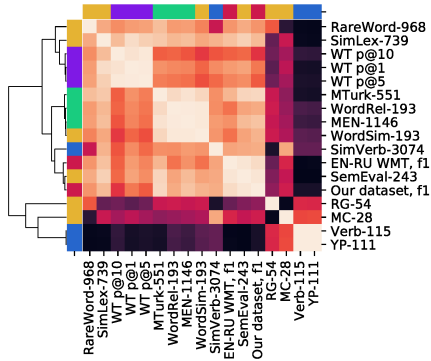}
    \captionsetup{justification=centering}
    \caption {Clustermap of different evaluation techniques. Lighter color correspond to stronger positive correlation. Each row and column is labeled according to benchmark type: red -- extrinsic, blue -- verbs, purple -- word translation, green -- word relatedness, yellow -- word similarity.}
\end{figure}

The results of the experiments with intrinsic and extrinsic evaluation are presented in Table 1. Despite the difference in scores for different models in one dataset could be minuscule, the scores for different intrinsic datasets vary a lot, and models that achieve higher results on one task often have lower results on other tasks. 

Figure 1 shows mutual similarities between datasets (measured as Spearman's rank correlation between evaluation scores from Table 1). One can see that there are at least 4 clusters: extrinsic+SemEval; word relations; word translation+some word similarities; others. 

Interestingly, \textit{SemEval} behaves similarly to extrinsic tasks: this benchmark contains not only single words but also two-word expressions (e.g. \textit{Borussia Dortmund}), so evaluation on this dataset is more similar to paraphrase detection task. Surprisingly, other word similarity datasets yield very different metrics. This is kind of unexpected, because paraphrase detection task relies on similarity of word senses.

Notably, many datasets from the same group (marked using color in the leftmost column on Figure 1) have difference in models' behavior (e.g. \textit{SimLex} and \textit{WordSim} both being word similarity benchmarks are clustered away from each other).

Our datasets, aligned models and code to reproduce the experiments are available at our GitHub~\footnote{\url{https://github.com/bakarov/cross-lang-embeddings}}.

\section{Conclusions and Future Work}

In this work we explored primary limitations of evaluation methods of intrinsic cross-language word embeddings. We proposed experiments on 5 models in order to answer the question `could we somehow estimate extrinsic performance of cross-language embeddings given some intrinsic metrics?'. Currently, the short answer is `No', but the longer is `maybe yes, if we understand the cognitive and linguistic regularities that take place in the benchmarks we use. Our point is that we not only need intrinsic datasets of different types if we want to robustly predict the performance of different extrinsic tasks, but we also should overthink the design and capabilities of existing extrinsic benchmarks.

Our research does not address some evaluation methods (like \textit{MultiQVEC} \cite{ammar2016massively}) and word embeddings models (for instance, \textit{Bivec} \cite{Luong-etal:naacl15:bivec}) since Russian do not have enough linguistic resources: there are certain parallel corpora available at \path{http://opus.nlpl.eu}, but a merge of all English-Russian  corpora has 773.0M/710.5M tokens, while the monolingual Russian model that we used in this study was trained on Wikipedia of 5B tokens (and English Wikipedia has a triple of this size). \textit{A fortiori}, these corpora have different nature (subtitles, corpus of Europar speeches, etc), and we think that merging them would yield a dataset of unpredictable quality.

In future we plan to make a comparison with other languages giving more insights about performance of compared models. We also plan to investigate cross-language extensions of other intrinsic monolingual tasks (like the analogical reasoning task) to make our findings more generalizable.    

\section*{Acknowledgments}

The reported study was funded by the Russian Foundation for Basic Research project 16-37-60048 mol\_a\_dk and by the Ministry of Education and Science of the Russian Federation
(grant 14.756.31.0001).

\bibliographystyle{acl_natbib}
\bibliography{naaclhlt2018}

\end{document}